\pdfoutput=1

\documentclass[11pt]{article}

\usepackage[preprint]{acl}

\usepackage{times}
\usepackage{latexsym}

\usepackage[T1]{fontenc}

\usepackage[utf8]{inputenc}

\usepackage{microtype}

\usepackage{inconsolata}

\usepackage{graphicx}
\usepackage{amsmath}

\title{Olapa-MCoT: Enhancing the Chinese Mathematical Reasoning Capability of LLMs}

\author{Shaojie Zhu$^{1}$ \space\space Zhaobin Wang$^{2}$ \space\space Chengxiang Zhuo$^{1}$ \space\space Hui Lu$^{2}$ \space\space Bo Hu$^{1}$ \space\space Zang Li$^{1}$\\
    $^{1}$Tencent \space\space\space\space $^{2}$Shanghai Jiao Tong University \\
    \{zsj\_alisy\}@sina.com \\
    \{wangzhaobin,huilu\}@sjtu.edu.cn \\
    \{felixzhuo,harryyfhu,gavinzli\}@tencent.com}

\begin{document}
\maketitle
\begin{abstract}
In the past two years, the outstanding performance of ChatGPT in multilingual and multitasking has led to large language models (LLMs) attracting widespread attention. However, restricted by expensive costs, many studies have to focus on the ability of only one major language. How can we quickly improve the model's capabilities in new languages without reducing its original capabilities under limited data and computing power? In this work, we focus on improving the Chinese mathematical reasoning capability based on Llama-2-13B, which is weak in Chinese mathematical reasoning. we proposed the Mathematical Chain of Thought method (Olapa-MCoT). First, we propose Similarity RRHF (SimRRHF), which adds the constraint of model optimization direction by introducing similarity loss based on RRHF. Furthermore, the novelty Incorrect Data Relearning (IDRL) method is designed, which improves the model’s ability to learn difficult knowledge. The experiment achieves significant performance, with the accuracy of Chinese mathematical reasoning up to 50\%, a 36\% rise compared to Llama-2-13B-chat. In addition, the accuracy of English reasoning ability also increased by nearly 4\%. It is worth mentioning that our method can be applied to any language major LLMs.
\end{abstract}

\section{Introduction}
Although the large language models (LLMs) have achieved state-of-the-art performance on various benchmarks, the performance in complex natural language processing (NLP) tasks is still not satisfactory. A typical task is mathematical reasoning \citep{cobbe2021training,lightman2023let}, which requires understanding math concepts, rigorous mathematical reasoning logic, and precise calculations. Therefore, it is still full of challenges to enhance the mathematical reasoning ability of LLMs. GPT-4 \citep{achiam2023gpt} is state-of-the-art in Chinese and English mathematical reasoning tasks, but it is a closed-source model. Meta has released Llama-2 \citep{touvron2023llama2} with various parameter sizes, which is a model learned mainly from a large amount of English corpus and is weak in Chinese mathematical reasoning. So as WizardMath \citep{luo2023wizardmath}. Some Chinese LLMs, such as the Baichuan \citep{Baichuan-13B,yang2023baichuan} and ChatGLM \citep{ChatGLM2-6B,ChatGLM3-6B} series, have improved their performance in Chinese mathematical reasoning to a certain extent. However, the universal LLMs obtained by pretraining and alignment learning on a large amount of Chinese corpus have expensive training costs and are difficult for many research teams to replicate. In this work, we focus on enhancing the Chinese mathematical reasoning capability based on Llama-2-13B lacking Chinese mathematical reasoning with only 100,000 Chinese mathematical reasoning data automatically constructed (see Appendix~\ref{sec:appendixsftdata}) by us and some open-source data.

Recent studies have focused on improving the mathematical reasoning capabilities of LLMs through optimizing prompts, aggregating reasoning paths, and alignment training. \textbf{Optimizing prompts.} Zero-shot-CoT \citep{kojima2022large}, Manual-CoT \citep{wei2022chain}, Auto-CoT \citep{zhang2022automatic}, Least-to-Most prompting \citep{zhou2022least} and Progressive-Hint Prompting (PHP) \citep{zheng2023progressive} improve performance by optimizing prompts. However, the information provided by the prompt is insufficient and inflexible for complex mathematical problems. \textbf{Aggregating reasoning paths.} Self-consistency \citep{wang2022self}, The Rationale-augmented Ensembles framework \citep{wang2022rationale}, Rejection-sampling Fine-Tuning (RFT) \citep{yuan2023scaling}, and \citet{xie2023olagpt} improve performance by generating a diverse set of reasoning paths. It does not apply to general mathematical problems. \textbf{Alignment training.} Reinforcement Learning from Human Feedback (RLHF) \citep{christiano2017deep,ouyang2022training,stiennon2020learning} is used to improve model performance usually. Which grounded on the PPO algorithm \citep{schulman2017proximal}, facilitates the alignment of LLMs with human preferences. However, PPO exhibits excessive sensitivity to hyperparameters and necessitates the concurrent loading of four models during training, which results in unstable training and excessive memory usage. In response to these challenges, the Rank Responses to align Human Feedback (RRHF) \citep{yuan2023rrhf} paradigm emerged as an alternative to RLHF. RRHF assigns scores to responses generated through various sampling strategies, aligning them with human preferences using ranking and sft loss. In contrast to RLHF which loads four models, RRHF achieves comparable results with only one or two models, significantly reducing memory usage and simplifying training. Inspired by RRHF, we also enhance performance based on the variance of RRHF.

In this work, we propose the Mathematical Chain of Thought method (Olapa-MCoT) to enhance the Chinese mathematical reasoning capability based on Llama-2-13B, without reducing its English mathematical reasoning capability. Our method proposes similarity RRHF (SimRRHF) so as not to deviate from the best object by introducing similarity loss into RRHF, and Incorrect Data Relearning (IDRL) is designed to improve the model’s learning ability for difficult knowledge. It is more surprising that experiments have shown that continuously adding new data cannot improve the model's ability when the model learning is insufficient but may reduce its ability.

\section{Methods}
This section introduces similarity RRHF (SimRRHF) in 2.1 and Incorrect Data Relearning (IDRL) in 2.2.
\subsection{Similarity RRHF (SimRRHF)}
Compared to the complex implementation process of RLHF \citep{ouyang2022training} and the high cost of maintaining 4 models (policy model, value model, reward model, reference model) during alignment, RRHF \citep{yuan2023rrhf} achieves comparable performance while significantly reducing memory and simplifying training. Inspired by RRHF, we propose SimRRHF to avoid deviating from the best object by introducing similarity loss. The motivation is we find some questions (see Appendix~\ref{sec:appendixsim}) that can be inferred correctly, but incorrect after RRHF alignment learning. After exploring the RRHF method, it is reasonable to speculate that this is due to the two losses in RRHF being both token-level learning, which focuses more on the increasing co-occurrence probability of tokens but ignores the overall mathematical logic correctness issue. To alleviate this problem, we introduce similarity loss by calculating the semantic similarity between candidate reasoning and target reasoning steps. In this way, it can constrain the model optimization direction. During learning, we found that SimRRHF can optimize the model faster and more stably.

We define each query as $x$, and each query has $k$ responses $y_i, 1\le i\le k$, these responses contain multiple CoT results, and reward model scores for each $y_i$, defined as $R(x, y_i)=r_i$. Our goal is to align the finetuned output of the model with the top-rated response, without deviating from the constraints of the model's performance.

Firstly, we retain the ranking loss calculation method of RRHF. Namely, model $\pi$ scores for each response $y_i$:
\begin{equation}
    p_i = \frac{\sum_t \log P_\pi(y_{i,t}|x,y_{i,<t})}{\|y_i\|},
\end{equation}
where $p_i$ is conditional log probability (length-normalized) of $y_i$ under model $\pi$. We optimize this object by ranking loss:
\begin{equation}
    L_{rank} = \sum_{r_i<r_j}\max(0,p_i-p_j).
\end{equation}

Secondly, for the sft loss, we directly use the length-normalized $p_i$ value of the top-rated response:
\begin{equation}
    i' = \arg\max_i{r_i},
\end{equation}
\begin{equation}
    L_{sft} = -p_{i'},
\end{equation}
$i'$ represents the index of the top-rated response.

Finally, we introduce similarity loss to measure the semantic distance between the response generated by model $\pi$ and the top-rated response.
\begin{equation}
L_{similarity}=1-cosine(E(y_{i'}),E(y_\pi)),
\end{equation}
where $E(y_{i'})$ is the mean embedding of the best response in samples, and $E(y_\pi)$ represents the mean embedding of the response generated by model $\pi$.

The total loss of SimRRHF is defined as the weighted sum of these three losses, $\lambda_{rank}$, $\lambda_{sft}$ and $\lambda_{sim}$ represent the weight of ranking loss, sft loss, and similarity loss respectively.
\begin{equation}
    L = \lambda_{rank} L_{rank} + \lambda_{sft} L_{sft} + \lambda_{sim} L_{similarity}.
\end{equation}
\subsection{Incorrect Data Relearning (IDRL)}
Mathematical reasoning task involves knowledge and reasoning logic with different levels of difficulty. It’s a great challenge to learn difficult knowledge for humans and models. During human learning, they collect incorrect data, by repeatedly learning incorrect data, to understand knowledge and improve exam scores. Inspired by this method, we design IDRL to learn repeatedly during alignment. 

As shown in Figure~\ref{fig:alignprocess} in Appendix, the procedure of IDRL contains three steps. First, collect data where the model makes incorrect inferences in the training dataset, which forms an important part of the samples for the next learning round. Second, supplement this with new samples as a complete training dataset for the next learning round. Lastly, train SimRRHF to enhance the model’s ability to learn difficult knowledge.

\section{Experimental Setup}
\subsection{Datasets}
\textbf{Sft datasets.} We collect the math data from some released datasets and construct Chinese mathematical reasoning data based on Ape210K\citep{zhao2020ape210k}. the details are in Appendix~\ref{sec:appendixsftdata}.
\textbf{Alignment datasets.} Appendix~\ref{sec:appendixalidata} shows the sample construction method for alignment datasets. Each query contains two responses generated by GPT-4 \citep{achiam2023gpt} and ChatGLM2-6B \citep{ChatGLM2-6B} respectively and scored by humans. In the first training round, the dataset size is 1,000 (D1), all from the above construction method. In the subsequent rounds, the size of the dataset is also 1,000 (D2), while it includes incorrect data and new samples. The incorrect data comes from the training dataset in the previous round that was incorrectly generated by the model. In addition, to verify the effectiveness of IDRL, a comparative 1,000 dataset is constructed without incorrect data (D3).
\textbf{Evaluation datasets.} The Chinese mathematical reasoning evaluation dataset is sampled from the Ape210K test dataset, with a size of 100. The English evaluation dataset is AQUA\footnote{\url{https://huggingface.co/datasets/deepmind/aqua_rat/viewer/raw/test}}, with a size of 254.
\subsection{Model}
Due to the lack of Chinese mathematical reasoning ability in Llama-2-13B, we first use QLoRA \citep{dettmers2023qlora} to finetune it with sft datasets to have a certain level of Chinese ability before alignment learning, named \textbf{Olapa-MCoT-SFT}. Based on Olapa-MCoT-SFT and D1, use SimRRHF to align the model, named \textbf{Olapa-MCoT-SimRRHF}. Then use IDRL and D2 to align the second round, named \textbf{Olapa-MCoT}. The hyper-parameters settings and ablation conclusions are in Appendix~\ref{sec:appendixhyper}.
\subsection{Baseline Models}
We aim to enhance the Chinese mathematical reasoning capability based on Llama-2-13B, without reducing its English mathematical reasoning capability. 

We choose \textbf{Baichuan-13B}, \textbf{Baichuan2-13B}, \textbf{ChatGLM2-6B}, and \textbf{ChatGLM3-6B} for performance comparison, which are representative and advanced open-source models for Chinese mathematical reasoning. At the same time, we also compare the Chinese and English mathematical reasoning ability to \textbf{GPT-3.5 Turbo} and \textbf{GPT-4}. 

Further, in RRHF, the total loss is defined as the unweighted sum of the ranking loss and sft loss, while in our SimRRHF, the total loss is defined as the weighted sum of the ranking loss, sft loss, and newly introduced similarity loss. To accurately analyze the importance of weighted and similarity loss, we design the baseline \textbf{Olapa-MCoT-RRHF} by aligning Olapa-MCoT-SFT with RRHF and \textbf{Olapa-MCoT-RRHF-weighted} by aligning Olapa-MCoT-SFT with weighted RRHF. Then, we design \textbf{Olapa-MCoT-D3} which is finetuned based on Olapa-MCoT-SimRRHF to verify the IDRL effectiveness. 

Finally, to verify that our method does not reduce the English mathematical reasoning ability of the model, we choose \textbf{Llama-2-13B} and \textbf{Llama-2-13B-chat} as the baseline.

\section{Results}
\subsection{Accuracy of Mathematical Reasoning}
As shown in Table~\ref{tab:accres1}, the evaluation results of Olapa-MCoT and all baseline LLMs in Chinese and English test datasets. It can be seen the Chinese mathematical reasoning accuracy of our Olapa-MCoT is up to 50\% compared to the base model Llama-2-13B lacking Chinese reasoning ability, which is 36\% higher than the Llama-2-13B-chat. At the same time, it is 3\% to 12\% higher than the Baichuan and ChatGLM series, and even 1\% more than the accuracy of GPT-3.5 Turbo. For the English mathematical reasoning capability, the accuracy shows that our Olapa-MCoT can still be no less than the Baichuan and ChatGLM series. Surprisingly, compared to the Llama-2-13B-chat, the English mathematical reasoning accuracy of our model has also increased by nearly 4\%. Besides, the non-convergence rate is much lower than Baichuan-13B and Llama-2-13B, as shown in Appendix~\ref{sec:appendixncr}.
\begin{table}
    \centering
    \begin{tabular}{lcc}
    \hline
    \textbf{model} & \textbf{ZH\_acc} & \textbf{EN\_acc}\\
    \hline
    Llama-2-13B & 0 & -- \\
    Llama-2-13B-chat & 14.00\% & 22.44\% \\
    Baichuan-13B & 38.00\% & 22.44\% \\
    Baichuan2-13B & 43.00\% & 23.62\% \\
    ChatGLM2-6B & 40.00\% & \textbf{26.38\%} \\
    ChatGLM3-6B & 47.00\% & 21.65\% \\
    GPT-3.5 Turbo & 49.00\% & \textbf{29.92\%} \\
    GPT-4 & \textbf{68.00\%} & \textbf{66.14\%} \\
    Olapa-MCoT(ours) & \textbf{50.00\%} & \textbf{26.38\%} \\
    \hline
    \end{tabular}
    \caption{The evaluation results of Olapa-MCoT and all baseline LLMs in Chinese and English test datasets. The second column ZH\_acc is the accuracy for Chinese mathematical reasoning, and the third column EN\_acc is the accuracy for English mathematical reasoning.}
    \label{tab:accres1}
\end{table}
\subsection{Value of SimRRHF}
We explore the value of SimRRHF from both accuracy and stability perspectives. 
\begin{table}
    \centering
    \begin{tabular}{lcc}
    \hline
    \textbf{model} & \textbf{ZH\_acc} & \textbf{best\_ckpt}\\
    \hline
    Olapa-MCoT-RRHF & 43.00\% & 4000 \\
    Olapa-MCoT- & & \\
    RRHF-weighted & 44.00\% & 2100 \\
    Olapa-MCoT-SimRRHF & \textbf{46.00\%} & 2000 \\
    \hline
    \end{tabular}
    \caption{The evaluation results of alignment are based on Olapa-MCoT-SFT. ZH\_acc is the accuracy for Chinese mathematical reasoning, and the best\_ckpt is the checkpoint at which the model reaches convergence.}
    \label{tab:accsim1}
\end{table}

As shown in Table~\ref{tab:accsim1}, compared to RRHF, the Olapa-MCoT-RRHF-weighted by only introducing weighted, the accuracy increases by 1\%, while our SimRRHF containing both weighted and similarity loss can increase by 3\%. At the same time, the convergence speed of our model has doubled.

As shown in Table~\ref{tab:accsim2}, compared to Olapa-MCoT-RRHF-weighted, SimRRHF not only improves the accuracy but also makes the model converge to the best performance faster and more stably.
\begin{table}
    \centering
    \begin{tabular}{lll}
    \hline
    \textbf{model} & \textbf{mean} & \textbf{var*1k}\\
    \hline
    Olapa-MCoT-RRHF & 41.80\% & 0.216 \\
    Olapa-MCoT- & & \\
    RRHF-weighted & 40.40\% & 0.904 \\
    Olapa-MCoT-SimRRHF & \textbf{43.20\%} & 0.456 \\
    \hline
    \end{tabular}
    \caption{The mean and thousand-fold variance of 5 saved ckpts near the best ckpt. we use mean and thousand-fold variance of accuracy to measure the model speed and stability.}
    \label{tab:accsim2}
\end{table}

\subsection{Effectiveness of IDRL}
As shown in Table~\ref{tab:accidrl}, our Olapa-MCoT by introducing IDRL, achieves an accuracy improvement of 4\%, while adding a new dataset D3 results in a 1\% decrease in accuracy. This is enough to prove that IDRL can enhance the model's ability to learn difficult knowledge, more exciting, it shows that continuously adding new data cannot improve the model's ability when the model learning is insufficient but may reduce its ability.
\begin{table}
    \centering
    \begin{tabular}{lc}
    \hline
    \textbf{model} & \textbf{ZH\_acc} \\
    \hline
    Olapa-MCoT-SimRRHF & 46.00\% \\
    Olapa-MCoT-D3 & 45.00\% \\
    Olapa-MCoT(ours) & \textbf{50.00\%} \\
    \hline
    \end{tabular}
    \caption{The evaluation results of alignment are based on Olapa-MCoT-SimRRHF. The second column ZH\_acc is the accuracy for Chinese mathematical reasoning.}
    \label{tab:accidrl}
\end{table}
\section{Conclusion}
In this work, we propose the Olapa-MCoT to enhance the Chinese mathematical reasoning capability based on Llama-2-13B, without reducing its English ability. Inspired by RRHF, we propose SimRRHF to avoid deviating from the best object by introducing similarity loss into RRHF and provide a novel IDRL learning method to improve the model’s learning ability for difficult knowledge. Experimental results show that our method consistently outperforms our provided baseline LLMs.
\section*{Limitations}
Although our study is based on the English ability model to enhance Chinese mathematical reasoning ability, during the experimental process, except for the corpus is language-related, the experimental methods and parameters are not limited by language. In theory, our method can be transferred to any language model, but the experimental verification will be carried out in the future. In addition, our research is based on language models and does not apply to models of other modalities.
\bibliography{oalpa_mcot,custom}

@article{wei2022chain,
  title={Chain-of-thought prompting elicits reasoning in large language models},
  author={Wei, Jason and Wang, Xuezhi and Schuurmans, Dale and Bosma, Maarten and Xia, Fei and Chi, Ed and Le, Quoc V and Zhou, Denny and others},
  journal={Advances in Neural Information Processing Systems},
  volume={35},
  pages={24824--24837},
  year={2022}
}

@article{kojima2022large,
  title={Large language models are zero-shot reasoners},
  author={Kojima, Takeshi and Gu, Shixiang Shane and Reid, Machel and Matsuo, Yutaka and Iwasawa, Yusuke},
  journal={Advances in neural information processing systems},
  volume={35},
  pages={22199--22213},
  year={2022}
}

@article{zhang2022automatic,
  title={Automatic chain of thought prompting in large language models},
  author={Zhang, Zhuosheng and Zhang, Aston and Li, Mu and Smola, Alex},
  journal={arXiv preprint arXiv:2210.03493},
  year={2022}
}

@article{zhou2022least,
  title={Least-to-most prompting enables complex reasoning in large language models},
  author={Zhou, Denny and Sch{\"a}rli, Nathanael and Hou, Le and Wei, Jason and Scales, Nathan and Wang, Xuezhi and Schuurmans, Dale and Cui, Claire and Bousquet, Olivier and Le, Quoc and others},
  journal={arXiv preprint arXiv:2205.10625},
  year={2022}
}

@article{zheng2023progressive,
  title={Progressive-hint prompting improves reasoning in large language models},
  author={Zheng, Chuanyang and Liu, Zhengying and Xie, Enze and Li, Zhenguo and Li, Yu},
  journal={arXiv preprint arXiv:2304.09797},
  year={2023}
}

@article{wang2022self,
  title={Self-consistency improves chain of thought reasoning in language models},
  author={Wang, Xuezhi and Wei, Jason and Schuurmans, Dale and Le, Quoc and Chi, Ed and Narang, Sharan and Chowdhery, Aakanksha and Zhou, Denny},
  journal={arXiv preprint arXiv:2203.11171},
  year={2022}
}

@article{wang2022rationale,
  title={Rationale-augmented ensembles in language models},
  author={Wang, Xuezhi and Wei, Jason and Schuurmans, Dale and Le, Quoc and Chi, Ed and Zhou, Denny},
  journal={arXiv preprint arXiv:2207.00747},
  year={2022}
}

@article{yuan2023scaling,
  title={Scaling relationship on learning mathematical reasoning with large language models},
  author={Yuan, Zheng and Yuan, Hongyi and Li, Chengpeng and Dong, Guanting and Tan, Chuanqi and Zhou, Chang},
  journal={arXiv preprint arXiv:2308.01825},
  year={2023}
}

@article{ouyang2022training,
  title={Training language models to follow instructions with human feedback},
  author={Ouyang, Long and Wu, Jeffrey and Jiang, Xu and Almeida, Diogo and Wainwright, Carroll and Mishkin, Pamela and Zhang, Chong and Agarwal, Sandhini and Slama, Katarina and Ray, Alex and others},
  journal={Advances in Neural Information Processing Systems},
  volume={35},
  pages={27730--27744},
  year={2022}
}

@article{christiano2017deep,
  title={Deep reinforcement learning from human preferences},
  author={Christiano, Paul F and Leike, Jan and Brown, Tom and Martic, Miljan and Legg, Shane and Amodei, Dario},
  journal={Advances in neural information processing systems},
  volume={30},
  year={2017}
}

@article{stiennon2020learning,
  title={Learning to summarize with human feedback},
  author={Stiennon, Nisan and Ouyang, Long and Wu, Jeffrey and Ziegler, Daniel and Lowe, Ryan and Voss, Chelsea and Radford, Alec and Amodei, Dario and Christiano, Paul F},
  journal={Advances in Neural Information Processing Systems},
  volume={33},
  pages={3008--3021},
  year={2020}
}

@article{lightman2023let,
  title={Let's Verify Step by Step},
  author={Lightman, Hunter and Kosaraju, Vineet and Burda, Yura and Edwards, Harri and Baker, Bowen and Lee, Teddy and Leike, Jan and Schulman, John and Sutskever, Ilya and Cobbe, Karl},
  journal={arXiv preprint arXiv:2305.20050},
  year={2023}
}

@article{luo2023wizardmath,
  title={Wizardmath: Empowering mathematical reasoning for large language models via reinforced evol-instruct},
  author={Luo, Haipeng and Sun, Qingfeng and Xu, Can and Zhao, Pu and Lou, Jianguang and Tao, Chongyang and Geng, Xiubo and Lin, Qingwei and Chen, Shifeng and Zhang, Dongmei},
  journal={arXiv preprint arXiv:2308.09583},
  year={2023}
}

@article{schulman2017proximal,
  title={Proximal policy optimization algorithms},
  author={Schulman, John and Wolski, Filip and Dhariwal, Prafulla and Radford, Alec and Klimov, Oleg},
  journal={arXiv preprint arXiv:1707.06347},
  year={2017}
}

@article{yuan2023rrhf,
  title={RRHF: Rank Responses to Align Language Models with Human Feedback without tears},
  author={Yuan, Zheng and Yuan, Hongyi and Tan, Chuanqi and Wang, Wei and Huang, Songfang and Huang, Fei},
  journal={arXiv e-prints},
  pages={arXiv--2304},
  year={2023}
}

@article{cobbe2021training,
  title={Training verifiers to solve math word problems},
  author={Cobbe, Karl and Kosaraju, Vineet and Bavarian, Mohammad and Chen, Mark and Jun, Heewoo and Kaiser, Lukasz and Plappert, Matthias and Tworek, Jerry and Hilton, Jacob and Nakano, Reiichiro and others},
  journal={arXiv preprint arXiv:2110.14168},
  year={2021}
}

@article{touvron2023llama2,
  title={Llama 2: Open foundation and fine-tuned chat models},
  author={Touvron, Hugo and Martin, Louis and Stone, Kevin and Albert, Peter and Almahairi, Amjad and Babaei, Yasmine and Bashlykov, Nikolay and Batra, Soumya and Bhargava, Prajjwal and Bhosale, Shruti and others},
  journal={arXiv preprint arXiv:2307.09288},
  year={2023}
}

@misc{Baichuan-13B,
    title = {A 13B large language model developed by Baichuan Intelligent Technology},
    author={Baichuan},
    year = {2023},
    howpublished = {\url{https://github.com/baichuan-inc/Baichuan-13B}}
}

@article{yang2023baichuan,
  title={Baichuan 2: Open large-scale language models},
  author={Yang, Aiyuan and Xiao, Bin and Wang, Bingning and Zhang, Borong and Bian, Ce and Yin, Chao and Lv, Chenxu and Pan, Da and Wang, Dian and Yan, Dong and others},
  journal={arXiv preprint arXiv:2309.10305},
  year={2023}
}

@misc{ChatGLM2-6B,
    title = {ChatGLM2-6B: An Open Bilingual Chat LLM},
    author = {ChatGLM2},
    year = {2023},
    howpublished = {\url{https://github.com/THUDM/ChatGLM2-6B}}
}

@misc{ChatGLM3-6B,
    title = {ChatGLM3 series: Open Bilingual Chat LLMs},
    author = {ChatGLM3},
    year = {2023},
    howpublished = {\url{https://github.com/THUDM/ChatGLM3}}
}

@misc{ji2023belle,
  title={Belle: Be everyone’s large language model engine},
  author={Ji, Yunjie and Deng, Yong and Gong, Yan and Peng, Yiping and Niu, Qiang and Ma, Baochang and Li, Xiangang},
  year={2023}
}

@inproceedings{abdullah2022chatgpt,
  title={ChatGPT: Fundamentals, applications and social impacts},
  author={Abdullah, Malak and Madain, Alia and Jararweh, Yaser},
  booktitle={2022 Ninth International Conference on Social Networks Analysis, Management and Security (SNAMS)},
  pages={1--8},
  year={2022},
  organization={IEEE}
}

@article{achiam2023gpt,
  title={Gpt-4 technical report},
  author={Achiam, Josh and Adler, Steven and Agarwal, Sandhini and Ahmad, Lama and Akkaya, Ilge and Aleman, Florencia Leoni and Almeida, Diogo and Altenschmidt, Janko and Altman, Sam and Anadkat, Shyamal and others},
  journal={arXiv preprint arXiv:2303.08774},
  year={2023}
}

@article{dettmers2023qlora,
  title={Qlora: Efficient finetuning of quantized llms},
  author={Dettmers, Tim and Pagnoni, Artidoro and Holtzman, Ari and Zettlemoyer, Luke},
  journal={arXiv preprint arXiv:2305.14314},
  year={2023}
}

@misc{xie2023olagpt,
      title={OlaGPT: Empowering LLMs With Human-like Problem-Solving Abilities}, 
      author={Yuanzhen Xie and Tao Xie and Mingxiong Lin and WenTao Wei and Chenglin Li and Beibei Kong and Lei Chen and Chengxiang Zhuo and Bo Hu and Zang Li},
      year={2023},
      eprint={2305.16334},
      archivePrefix={arXiv},
      primaryClass={cs.CL}
}

@article{zhao2020ape210k,
  title={Ape210k: A large-scale and template-rich dataset of math word problems},
  author={Zhao, Wei and Shang, Mingyue and Liu, Yang and Wang, Liang and Liu, Jingming},
  journal={arXiv preprint arXiv:2009.11506},
  year={2020}
}

\appendix

\section{Datasets}
\subsection{Sft Datasets}
\label{sec:appendixsftdata}
\begin{figure}[t]
    \includegraphics[width=\columnwidth]{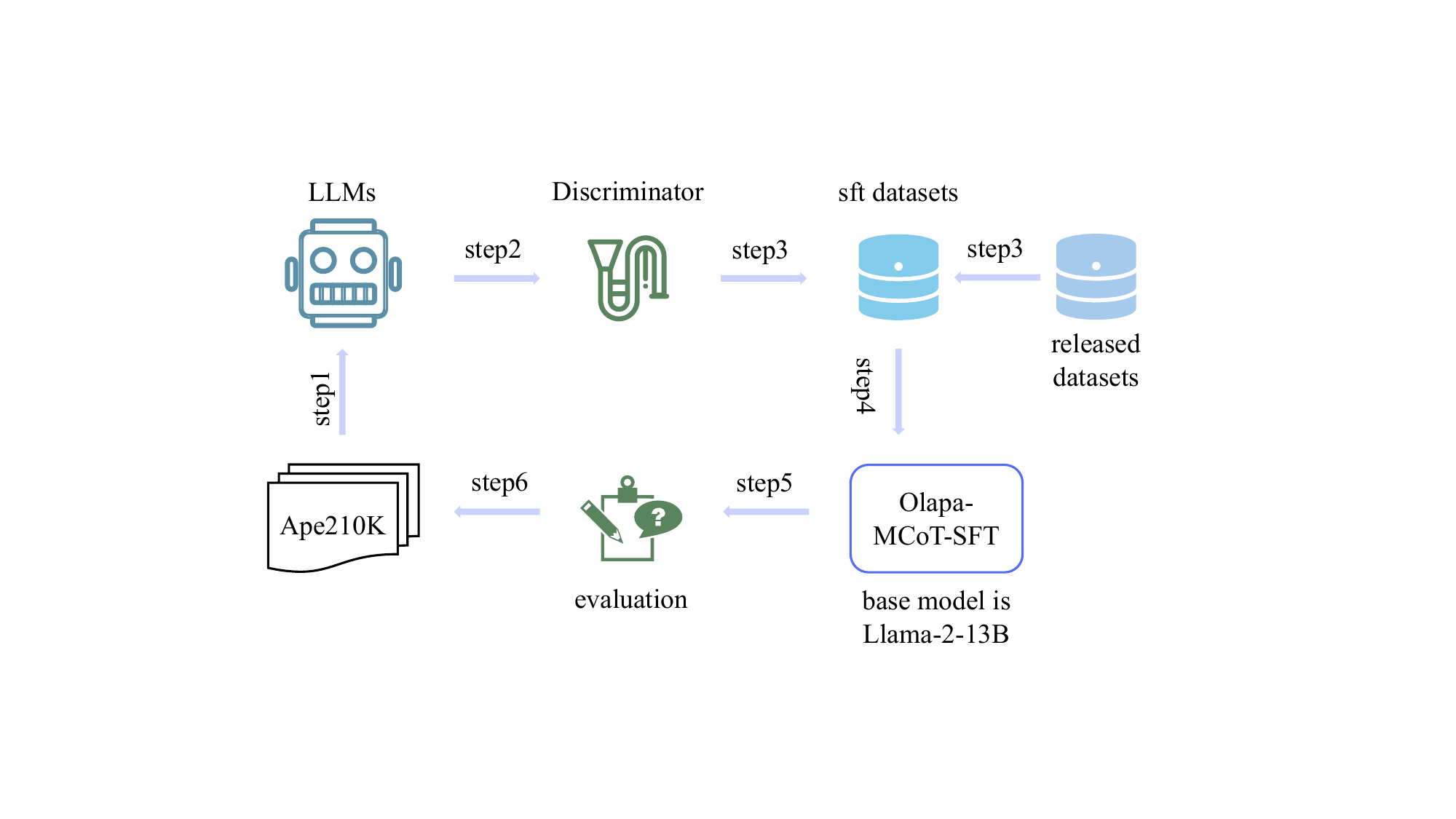}
    \caption{The overall process of sft. Steps 1 to 3 describe the process of collecting sft data, step 4 is sft training, and step 5 to 6 describe whether to continue training based on the accuracy of the model.}
    \label{fig:sftprocess}
\end{figure}

\paragraph{Auto-Constructed by LLMs.} High-quality Chinese mathematical reasoning samples are very scarce. To improve the diversity and quality of sft datasets, we use open-source LLMs based on Ape210K to automatically construct a batch of high-quality samples. In addition, we collect the math data from the denoised BELLE released about 237K\footnote{\url{https://huggingface.co/datasets/BelleGroup/school_math_0.25M}}, the original English data of OpenAI released PRM\footnote{\url{https://github.com/openai/prm800k/tree/main}}, about 30K based on the Chinese translation results by GPT-3.5 Turbo, and less than 1K program data released by firefly-train-1.1M\footnote{\url{https://huggingface.co/datasets/YeungNLP/firefly-train-1.1M}}. The details of the construction method and samples are in Appendix~\ref{sec:appendixsftdata}.
The process of constructing is shown from step 1 to step 3 in Figure~\ref{fig:sftprocess}. As follows:

Step 1: we design 20 instructions in total, while each prompt consists of a question and an instruction, for example, the question is sampled from Ape210K, and the instruction is “Let me think step by step”.

Step 2: the reasoning steps are generated by open-source LLMs, in our work, they are Baichuan-13B and ChatGLM2-6B. Given a prompt, the model outputs the reasoning steps and its answer.

Step 3: we design a high-quality discriminator to filter out samples with correct answers and better reasoning steps. The discriminator is achieved by length limitation within 800 and the output answer is equal to the referred answer in Ape210K, the accuracy is over 95\%.
\paragraph{Distribution.} The distribution of sft datasets is shown in Table~\ref{tab:sftdata}. A total of approximately 367K MCoT samples were collected.
\begin{table*}
    \centering
    \begin{tabular}{lccc}
    \hline
    \textbf{name} & \textbf{source} & \textbf{size} & \textbf{language} \\
    \hline
    olapa.zh.cot.train.json & ape210K + LLMs & 96,594 & Chinese  \\
    school.math.clean.json & BELLE & 237,349 & Chinese \\
    prm.en.ground.truth.json & OpenAI & 10,553 & English \\
    prm.en.generator.json & OpenAI & 11,624 & English \\
    prm.zh.generator.json & OpenAI + translate & 9,590 & Chinese \\
    program.firefly.json & Firefly & 974 & Chinese + English \\
    \hline 
    \textbf{total} & & \textbf{366,684} & \textbf{Chinese + English} \\
    \hline
    \end{tabular}
    \caption{The distribution of sft datasets.}
    \label{tab:sftdata}
\end{table*}
\subsection{Alignment Datasets}
\label{sec:appendixalidata}.
\begin{figure}[t]
    \includegraphics[width=\columnwidth]{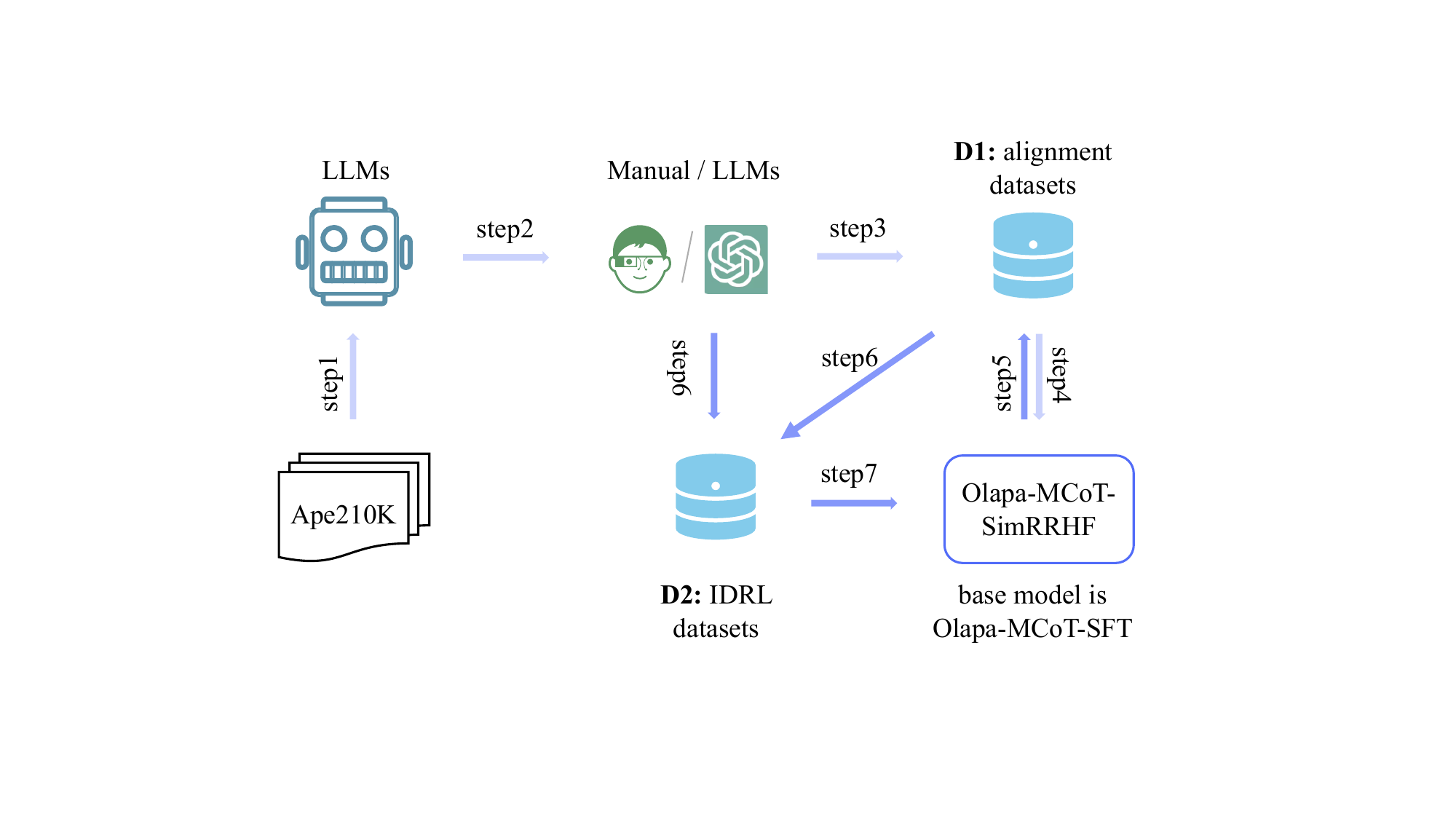}
    \caption{The overall process of alignment. Step 1 to 3 describe the process of conducting alignment data, step 4 is alignment training, and step 5 to 7 describe the process of collecting the IDRL datasets.}
    \label{fig:alignprocess}
\end{figure}
\paragraph{Auto-Constructed by LLMs.} The process is shown from step 1 to step 3 in Figure~\ref{fig:alignprocess}. As follows:

Step 1: we use Chinese instruction, while each prompt consists of a question and an instruction, the question is also sampled from Ape210K, and the Chinese instruction is translated from “Let me think step by step”.

Step 2: the reasoning steps are generated by open-source LLMs, in our work, they are GPT-4 and ChatGLM2-6B. Given a prompt, the model outputs the reasoning steps and its answer.

Step 3: merge the model outputs based on the same question, and after merging, there are two candidate responses for each question. Manually sort the candidates.

\section{SimRRHF Cases}
\label{sec:appendixsim}
As shown in Table~\ref{tab:badcase}, they are sample bad cases after alignment training by RRHF, in the table, they are translated from Chinese.
\begin{table*}
    \centering
    \begin{tabular}{p{4cm} p{4cm} p{4cm}}
    \hline
    \textbf{QA\_pair} & \textbf{result1} &\textbf{result2} \\
    \hline
    \textbf{Q:} What is the quotient of the sum of 45 and 39 divided by their difference? What is the correct formula? \textbf{A:} 5.25. (45+39) / (45-39). & Firstly, we need to calculate the sum and difference between 45 and 39. And: 45+39=84. Difference: 45-39=16. Next, we need to divide the sum by the difference. Quotient: 84 / 16=5.25. \color{blue}\textit{So, the sum of 45 and 39 divided by their difference gives a quotient of 5.25. The correct formula is (45+39) / (45-39) =84 / 16=5.25.} & Firstly, we need to calculate the sum of 45 and 39 and their difference. And: 45+39=84. Difference: 45-39=16. Then, we divide these two numbers to obtain the quotient. So, dividing 84 by 16 equals 5 over 11. \color{red}\textit{So, the correct formula should be 5/11.} \\
    \hline
    \textbf{Q:} The two brothers collected Warcraft cards together. The younger brother collected the number of cards equal to \textit{a}, and the number of cards collected by the older brother was his (5/3). How many cards did the two brothers collect in total? \textbf{A:} (8/3)\textit{a}. & We can solve this problem by following these steps: Firstly, we know that our younger brother has collected the number of cards equal to \textit{a}. Secondly, the number of Warcraft cards collected by the older brother is 5/3 of his own. So, my brother collected \textit{a} * (5/3) = (5/3)\textit{a} Warcraft cards. Finally, we added up the number of Warcraft cards collected by our younger brother and brother to obtain a total of \textit{a} + (5/3)\textit{a} = (8/3)\textit{a} Warcraft cards collected by the two brothers. \color{blue}\textit{So, the two brothers collected a total of (8/3)\textit{a} Warcraft cards.} & Firstly, we know that our younger brother collected the number of cards equal to \textit{a}, so what he collected is \textit{a} * (5/3) = 2.5\textit{a}. \color{red}\textit{So, the two brothers collected a total of \textit{a} + 2.5\textit{a} = 3.5\textit{a}.} \\
    \hline
    \end{tabular}
    \caption{The bad cases in alignment training with RRHF. QA\_pair contains the question which is the model input and the refer answer. The Result1 column is the model’s correct output before alignment training, and the result2 column is the model’s incorrect output after alignment training.}
    \label{tab:badcase}
\end{table*}
\section{Hyper-parameters}
\label{sec:appendixhyper}
\textbf{SFT hyper-parameters.} We set the $max\_length$ as 1024, $batch\_size$ as 32,  $learning\_rate$ as 1e-4, the $r$ value of QLoRA as 8, and $alpha$ as 16.
\textbf{Alignment hyper-parameters.} We set the $max\_length$ as 1024, $batch\_size$ as 4, $alpha$ of QLoRA as 16, $learning\_rate$ as 1e-5. Training SimRRHF with $\lambda_{rank}$ as 10.0, $\lambda_{sft}$ as 1.0, $\lambda_{sim}$ as 1.0. Training Olapa-MCoT-RawRRHF-weighted with $\lambda_{rank}$ as 10.0, and $\lambda_{sft}$ as 1.0. We conducted an ablation study, and the weight of similarity loss set 1.0, 2.0, and 5.0, the result shows that the accuracy has slightly decreased by about 1\% when the weight is 2.0 and 5.0.

\section{Additional Experimental Results}
\subsection{Non-convergence Rate}
\label{sec:appendixncr}
The non-convergence rate of LLMs is shown in Table~\ref{tab:ncrres1}.
\begin{table*}
    \centering
    \begin{tabular}{lcc}
    \hline
    \textbf{model} & \textbf{ZH\_Non-convergence Rate} & \textbf{EN\_Non-convergence Rate}\\
    \hline
    Llama-2-13B & 30.00\% & -- \\
    Llama-2-13B-chat & 11.00\% & 0 \\
    Baichuan-13B & 5.00\% & 10.24\% \\
    Baichuan2-13B & 3.00\% & 1.57\% \\
    ChatGLM2-6B & 1.00\% & 1.18\% \\
    ChatGLM3-6B & 1.00\% & 0 \\
    GPT-3.5 Turbo & 0 & 0 \\
    GPT-4 & 0 & 0.39\% \\
    Olapa-MCoT(ours) & 1.00\% & 1.57\% \\
    \hline
    \end{tabular}
    \caption{The Non-convergence Rate of Olapa-MCoT and all baseline LLMs in Chinese and English test datasets. The second column ZH\_Non-convergence Rate is the value for Chinese mathematical reasoning, and the third column EN\_Non-convergence Rate is the value for English mathematical reasoning. Non-convergence is defined as the situation where the output has a loop or does not end within a finite length.}
    \label{tab:ncrres1}
\end{table*}

\end{document}